\documentclass[letterpaper, 10 pt, conference]{ieeeconf}  

\IEEEoverridecommandlockouts                              

\usepackage{graphicx}

\usepackage[caption=false]{subfig}
\usepackage{xcolor}
\usepackage{array}
\usepackage{float}
\usepackage{booktabs}
\usepackage{adjustbox}
\usepackage[noadjust]{cite}
\usepackage{multirow}
\usepackage{todonotes}
\usepackage{marginnote}

\usepackage[a-1b]{pdfx}

\title{\LARGE \bf
Analysis of Forces Exerted by Shoulder and Elbow Fabric-based Pneumatic Actuators for Pediatric Exosuits}

\author{Mehrnoosh Ayazi,$^{1}$ Ipsita Sahin,$^{2}$ Caio Mucchiani,$^{1}$ Elena Kokkoni,$^{2}$ and Konstantinos Karydis$^{1}$ 
\thanks{$^{1}$~Dept. of Electrical and Computer Engineering; $^{2}$~Dept. of Bioengineering, University of California, Riverside, 900 University Ave, Riverside, CA 92521, USA. Email:{\tt\footnotesize\{mayaz004, isahi001, caiocesr, elenak, karydis\}@ucr.edu}. 
We gratefully acknowledge the support of NSF \# CMMI-2133084. 
Any opinions, findings, and conclusions or recommendations expressed in this material are those of the authors and do not necessarily reflect the views of the National Science Foundation.
}}

\begin{document}
\maketitle
\thispagestyle{empty}
\pagestyle{empty}

\begin{abstract}
To enhance pediatric exosuit design, it is crucial to assess the actuator-generated forces. This work evaluates the contact forces exerted by soft fabric-based pneumatic actuators in an upper extremity pediatric exosuit. Two actuators were examined: a single-cell bidirectional actuator for shoulder abduction/adduction and a bellow-type actuator for elbow extension/flexion. Experiments assessed the impact of actuator anchoring points and the adjacent joint's angle on exerted forces and actuated joint range of motion (ROM). These were measured via load cells and encoders integrated into a custom infant-scale engineered apparatus with two degrees of freedom (two revolute joints).  For the shoulder actuator, results show that anchoring it further from the shoulder joint center while the elbow is flexed at $90^\circ$ yields the highest ROM while minimizing the peak force exerted on the body. For the elbow actuator, anchoring it symmetrically while the shoulder joint is at $0^\circ$ optimizes actuator performance. 
These findings contribute a key step toward co-optimizing the considered exosuit design for functionality and wearability.
\end{abstract}

\section{Introduction}
Soft exosuits have shown significant promise in assistance, rehabilitation, and human augmentation across diverse applications. 
Existing efforts have targeted lower limb mobility, such as in walking~\cite{banyarani2024design}, hip motion~\cite{thalman2021hip}, 
knee function~\cite{park2020hinge, sridar2017development}, and ankle support~\cite{bae2018lightweight}, as well as upper limb mobility, including assistance at the shoulder joint~\cite{Simpson2020, o2017soft}, elbow support~\cite{chiaradia2018design, thalman2018novel}, wrist support~\cite{schaffer2024soft}, and hand function~\cite{ge2020design}. 
The vast majority of these efforts consider adult users; limited attention to devices specifically designed for children, particularly those that assist upper extremity (UE) movement in very young children, has been provided so far~\cite{Li2019DesignExo, arnold2020exploring, Kokkoni2020_asme}. 
Given the potential of assistive technology to improve motor function~\cite{henderson2008assistive}, more research is needed in pediatric exosuits.

A key measure of exosuit performance is the force generation capability which has been examined mainly on devices aimed at adults~\cite{schiele2009influence, xiloyannis2019physiological, bhardwaj2023manipulating, georgarakis2018method, ito2019control}. 
Both the magnitude and direction of the exerted (contact) forces, as well as their point of application play a role.  
Excessive or improperly targeted amounts of force (or pressure) can restrict natural movement and cause skin problems, circulation problems, and injuries~\cite{mak2010biomechanics,bringard2006effects, mayrovitz2003effects, schiele2009influence}, ultimately resulting in discomfort and reduced acceptance of the device~\cite{bright2013assistive}. 
As such, detailed profiling of contact forces can provide insight into actuator limitations and dynamic behaviors which in turn can be used to improve the device and its underlying controllers (e.g.,~\cite{mucchiani2022closed, mucchiani2023robust}).
The \emph{desired} force generation capacity of an exosuit is typically determined by the intended task and type of motion assistance (e.g., whole-arm reaching, fine manipulation), thereby guiding component selection and system design. 
The \emph{achieved} force generation capacity is determined by multiple factors, crucially including among them the anchoring points of the actuators.

\begin{figure}[!t]
\vspace{6pt}
     \centering
     \includegraphics[width=0.9\columnwidth]{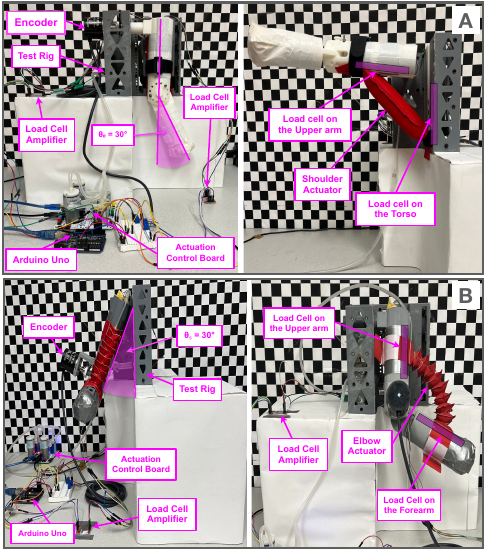}
     \vspace{-10pt}
         \caption{Experimental setup for force profiling of a fabric-based pneumatic (A) shoulder and (B) elbow actuator. The setup allows various actuator anchoring points and non-actuated joint angle settings (locked in place).}
     \label{fig:setup}
     \vspace{-21pt}
\end{figure}

The actuator anchoring points relative to the body can significantly influence joint kinematics and kinetics by facilitating force transfer to the body and stabilizing the body-exosuit interaction~\cite{Escudero2020, bae2018biomechanical, kokubu2024deriving, wehner2013lightweight}. 
Our previous work~\cite{sahin2024evaluation} assessed the effect of varying the anchoring points of several different pneumatic actuators on end-effector kinematics, enabling an informed selection of the actuators used herein. 
A single-cell inflatable actuator~\cite{sahin2022bidirectional} for the shoulder led to greater range of motion (ROM) and end-effector path length, while a 10-cell bellow-type actuator~\cite{sahin2023fabric} with square cells was more appropriate for the elbow in terms of maximizing ROM. 
In both cases, kinematic performance (i.e. ROM, motion smoothness, etc.) was improved when the anchoring points intersect along the mid-auxiliary line of the body~\cite{sahin2024evaluation}. 
However, the previous work did not assess joint kinetics and the exerted forces on the body as anchoring points vary.

The goal of this paper was to assess the exerted forces from the aforementioned actuators on the torso, upper arm, and forearm of an engineered apparatus based on 12-month-old infant anthropometrics (Fig.~\ref{fig:setup}). 
In addition, a preliminary assessment of the underlying relation between input pressure and exerted forces is provided for future automatic controller design.
Experiments involved one joint's actuator being active while the other joint remained locked in place at different points. 
We analyzed the exerted forces and the actuated joint's ROM under varying the active actuator anchoring points and the non-actuated joint's locked angle value. 
Results reveal optimal actuator anchoring points and the cases where exerted forces are minimized and ROM is maximized for the same actuator input pressure. 
Taken together with our previous findings~\cite{sahin2024evaluation}, this work provides key information about how to place actuators onto a soft pediatric exosuit to improve its function.

\section{Materials and Methods} 
\subsection{Experimental Setup}
Two types of soft, fabric-based pneumatic actuators were selected for their range of motion, ease of manufacturing, and durability.  
A single-cell rectangular actuator~\cite{sahin2022bidirectional} was used for shoulder abduction/adduction (achieved through inflation/deflation, respectively, with deflation-driven adduction due to reduced upper arm support).  
Elbow flexion/extension was facilitated via inflation and deflation, respectively, of a 10-cell bellow-type actuator~\cite{sahin2023fabric}. 

Two 3D-printed test rigs (Fig.~\ref{fig:setup}) were designed based on anthropometric data of 12-month-old infants in the 50th percentile~\cite{edmond2020normal, Fryar2021}. 
Each rig was configured to support a specific type of actuator (either shoulder or elbow) to study their individual force profiles. 
Both rigs consist of a base as the torso, an upper arm (UA), a forearm (FA), and a small part weighing $0.06$\;kg attached at the end of the FA in lieu of the hand. 
The UA ($16.4$\;cm in length, $14.7$\;cm in circumference) was connected to the torso via a revolute joint, effectively allowing for shoulder adduction and abduction. 
The FA ($10.85$\;cm in length and $14.51$\;cm in circumference) was attached to the UA through another revolute joint, allowing for elbow flexion and extension. 
Both UA and FA segments were hollow and filled in with sandbags to match a desired UE weight.  
The combined weight of UA and FA was set at $0.432$\;kg, based on anthropometrics of an infant of total body mass of $10.8$\;kg~\cite{zernicke1992mass}. 

The primary distinction between the two rigs lies in the positioning of sensors, which are used to measure the normal force generated by the actuator as well as the respective joint angle.\footnote{~By virtue of this work's actuator mechanical design and placement along the body's mix-auxiliary line, the normal force component is the one primarily contributing to mechanical work associated with the intended motion support as compared to a small but still present shear force component. For this reason, we prioritized the measurement and analysis of normal forces. The inclusion of shear forces is part of future work, as we delve into further motion support (e.g., shoulder flexion/extension).}
Each test rig incorporated two load cells (YZC-133) with a capacity of $5$\;kg to gauge the contact force exerted by the actuators on the UA, torso, or FA. 
The placement of these load cells was determined based on the impact area of the actuator during operation. 
The shoulder actuator exerts forces on the torso and UA. 
In the rig designed for the shoulder actuator, one load cell was situated within the UA, aligned with the vertical centerline on its inner side. 
This load cell was coupled with a plate measuring $11.1\times 1.36$\;cm to effectively capture the force applied over the area covered by the plate. 
The second load cell was positioned at the torso of the rig, also aligned with the vertical centerline on the inner side facing the UA, and was attached to a $15.6\times1.66$\;cm plate. 
Conversely, the elbow actuator exerts forces on the UA and FA.
In the respective test rig, the load cells were installed 
and aligned with the UA and FA vertical centerlines on the front side. 
The plates connected to these load cells measured $11.6\times1.66$\;cm for the UA and $8.85\times1.66$\;cm for the FA.
Each load cell was connected to an amplifier (HX711, Sparkfun) to convert the output data into a readable format at a frequency of $11$\;Hz, ultimately linked to the ADC pin of a microcontroller board (Arduino UNO). 
Additionally, a 1024 pulse per revolution rotary encoder (E6B2, Sparkfun) was integrated into the actuated joint to measure the respective joint angle. 
Encoder data were collected with the use of the microcontroller. 
The latter transmitted all collected data (amplified load cell and rotary encoder data) to a host computer for data logging and follow-on analysis. 

The inflation and deflation of the actuators were managed using an external pneumatic board (Programmable-Air kit). This board features two compressor/vacuum pumps and three pneumatic valves to control the airflow at a rate of $2$\;L/min. 
The airflow is adjusted through the pump duty cycle in $[0-100]$\%. 
The pressure can vary between $[-50,50]$\;kPa. 
The board is equipped with a pressure sensor (SMPP-03) and its own microcontroller (Arduino Nano). 
The integrated Arduino allows for connectivity to the host computer via a serial-to-USB interface, enabling users to send commands to the pumps and valves and to log sensor data.

\subsection{Experimental Protocol}
\begin{figure}[!t]
\vspace{4pt}
     \centering
     \includegraphics[width=0.75\columnwidth]{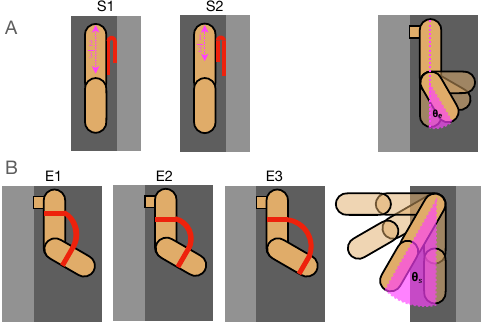}
     \vspace{-9pt}
         \caption{(A) Schematic diagram of shoulder actuator anchoring: Two anchoring points (S1 at two-thirds and S2 at one-half of the upper arm length from the proximal end) and four fixed elbow angles ($0^{\circ}$, $30^{\circ}$, $60^{\circ}$, and $90^{\circ}$). (B) Schematic diagram of elbow actuator anchoring: Twelve configurations with three anchoring points (E1, E2, E3) and four fixed shoulder angles ($0^{\circ}$, $30^{\circ}$, $60^{\circ}$, and $90^{\circ}$). E1: Actuator cells distributed 6:4 (UA:FA), with attachment points at two-thirds (UA) and one-half (FA) of their lengths. E2: Cells distributed 1:1, with both attachment points at one-half of their lengths. E3: Cells distributed 4:6, with attachment points at one-half (UA) and four-fifths (FA) of their lengths. Actuators shown in thick (red) curves.}
     \label{fig:conditions}
     \vspace{-15pt}
\end{figure}

A series of experiments were conducted to explore how two key elements affect joint ROM and force exertion of each actuator on the test rigs.
The first element pertains to variations in the anchoring of the actuators. 
Two anchoring configurations were tested for the shoulder actuator: one at two-thirds (S1) and the other at one-half (S2) of the UA length, measured from its proximal end (Fig.~\ref{fig:conditions}A left panel). 
The other end of the actuator was attached to the torso. 
Three anchoring configurations were tested for the elbow actuator based on the distribution of the actuator's cells along the UA and FA: one symmetric (E2) and two asymmetric configurations (E1 and E3). 
In the E1 configuration, more cells extended on the UA (about $60$\%) than on the FA, while the opposite was true in E3. 
In E2, there was a symmetrical distribution of cells on both segments (Fig.~\ref{fig:conditions}B left panel). 
The second element refers to the effect the non-actuated joint being locked in place at different points can have on the active joint's ROM and force exertion of the actuators. 
One of the test rigs allowed the elbow joint to be locked at four angles, $\theta_e\in\{0^{\circ},30^{\circ},60^{\circ},90^{\circ}\}$ (Fig.~\ref{fig:conditions}A right panel), while the other featured shoulder joint locking at four angles, $\theta_s\in\{0^{\circ},30^{\circ},60^{\circ},90^{\circ}\}$ (Fig.~\ref{fig:conditions}B right panel).

A total of 30 trials were performed for each configuration. 
Load cells and encoders integrated into the test rigs measured the exerted force and the angular displacement throughout each trial.
Further, input pressure was collected from the pressure sensor on the pneumatic board. 
Before each trial, the load cell was set to zero with the actuator anchored but not actuated (to ensure that measured forces resulted solely from the actuation during the trials). 
Each trial for the shoulder actuator lasted 10 sec, consisting of a 5-sec inflation phase followed by a 5-sec deflation phase. 
The elbow actuator trials also lasted 10 sec each, containing a 5-sec deflation phase followed by 5 sec of inflation. To enable these 5-sec actuation phases, the pumps operated at a $100$\% duty cycle.~\footnote{~Achieving rapid actuation is important since infant reaching at this age can be fast, typically lasting about two sec~\cite{zhou2021infant}. Here we used 5-sec phases to facilitate analysis. However, as we will show shortly, the actuators can reach maximum ROM within three and two sec for the shoulder and elbow respectively, thus making the exosuit appropriate for our target application.}

\subsection{Variables of Interest and Statistical Analysis} 

The computed variables were joint ROM (for both shoulder and elbow joints) and peak force exerted by each actuator on the relevant body segment. 
To assess how the actuator anchoring configurations and non-actuated locked joint angles affected these two variables, a series of separate non-parametric tests was performed for each factor and actuator (violation of normality was confirmed with the Kolmogorov-Smirnov test).
For the shoulder actuator, Mann-Whitney U tests were conducted to identify significant differences in shoulder joint ROM and peak force between the two shoulder actuator anchoring configurations (S1, S2).
Kruskal-Wallis H tests were performed to evaluate differences in shoulder joint ROM and peak force across the four non-actuated elbow joint locked angles {$\theta_e$}. 
For the elbow actuator, Kruskal-Wallis H tests were conducted to assess significant differences in elbow joint ROM and peak force across the elbow actuator anchoring configurations (E1, E2, E3).
Lastly, Kruskal-Wallis H tests were performed to evaluate differences in elbow joint ROM and peak force across the four non-actuated shoulder joint locked angles {$\theta_s$}.  
Post-hoc pairwise comparisons were conducted as needed.
The significance level at $0.05$ was Bonferroni-adjusted to account for multiple comparisons.
Statistical analyses were conducted with SPSS v.30.

\section{Results and Discussion} 
\subsection{Shoulder Actuator Assessment}
\begin{figure}[!t]
\vspace{6pt}
     \centering
     \includegraphics[trim={0.2cm, 7.5cm, 2cm, 2cm},clip,width=1\columnwidth]{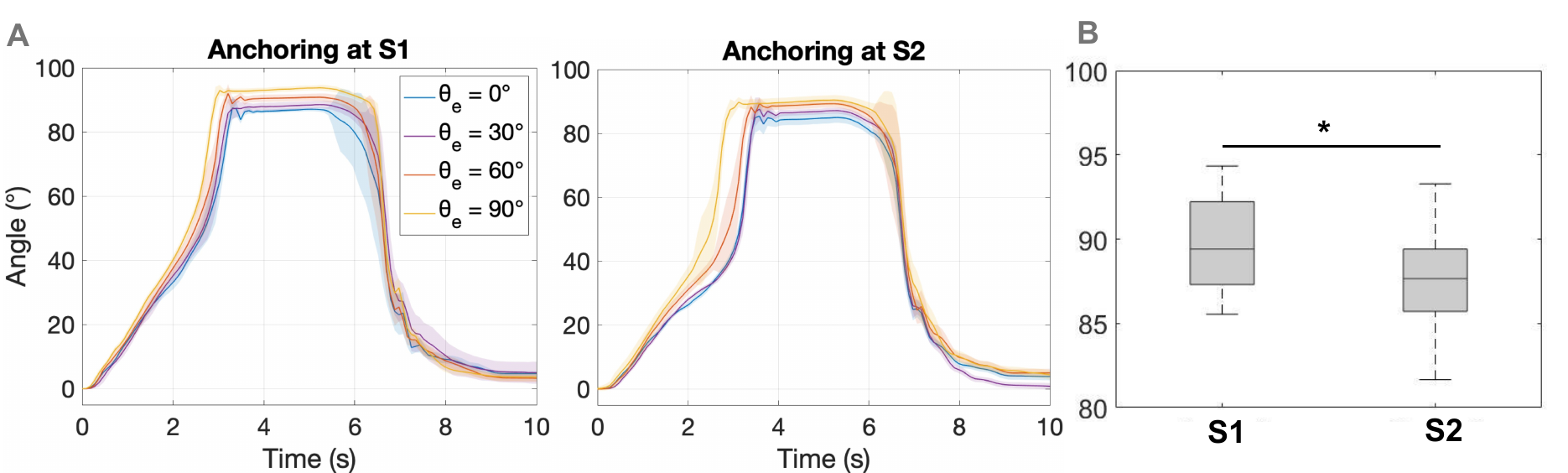}
     \vspace{-20pt}
         \caption{(A) Changes in shoulder joint angle over time in the two shoulder actuator anchorings for different locked elbow angles. Shaded areas represent one standard deviation. (Best viewed in color.) (B) Boxplot comparing the shoulder joint angle between the S1 and S2 configurations.} 
     \label{fig:shoulder_AngleTime}
     \vspace{-12pt}
\end{figure}

The change in shoulder joint angle throughout the trial and across the different non-actuated elbow joint angles for both shoulder actuator anchoring configurations is depicted in Fig.~\ref{fig:shoulder_AngleTime}A.
Overall, shoulder joint ROM was greater in S1 than S2 ($U=4184$, $p < 0.001$; Fig.~\ref{fig:shoulder_AngleTime}B).
There was an effect of non-actuated elbow joint locked angles on ROM for both S1 ($\chi\textsuperscript{2}(3)=102.281$, $p<0.001$) and S2 ($\chi\textsuperscript{2}(3)=83.999$, $p<0.001$) configurations. 
Post-hoc tests demonstrated that locking the elbow angle at $60^{\circ}$ and $90^{\circ}$ leads to a greater shoulder joint ROM than being locked at $0^{\circ}$ and $30^{\circ}$ for both anchoring configurations ($p < 0.001$ for all comparisons).

\begin{figure}[!t]
\vspace{6pt}
     \centering
\includegraphics[width=0.95\columnwidth]{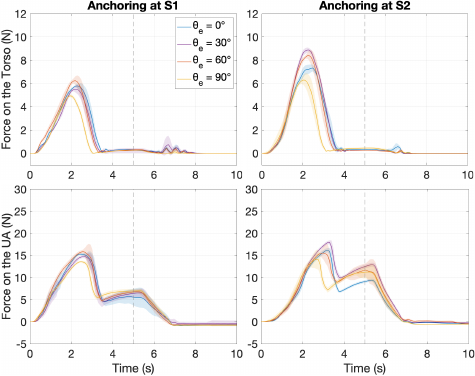}
     \vspace{-6pt}
         \caption{Temporal evolution of the forces exerted by the shoulder actuator on the torso (Top) and UA (Bottom) for the two shoulder actuator anchorings and different locked elbow angles.
         }
     \label{fig:force vs. time}
     \vspace{-16pt}

\end{figure}
\begin{figure}[!t]
\vspace{3pt}
     \centering
\includegraphics[trim={1.5cm, 2cm, 6cm, 2cm},clip, width=0.95\columnwidth]{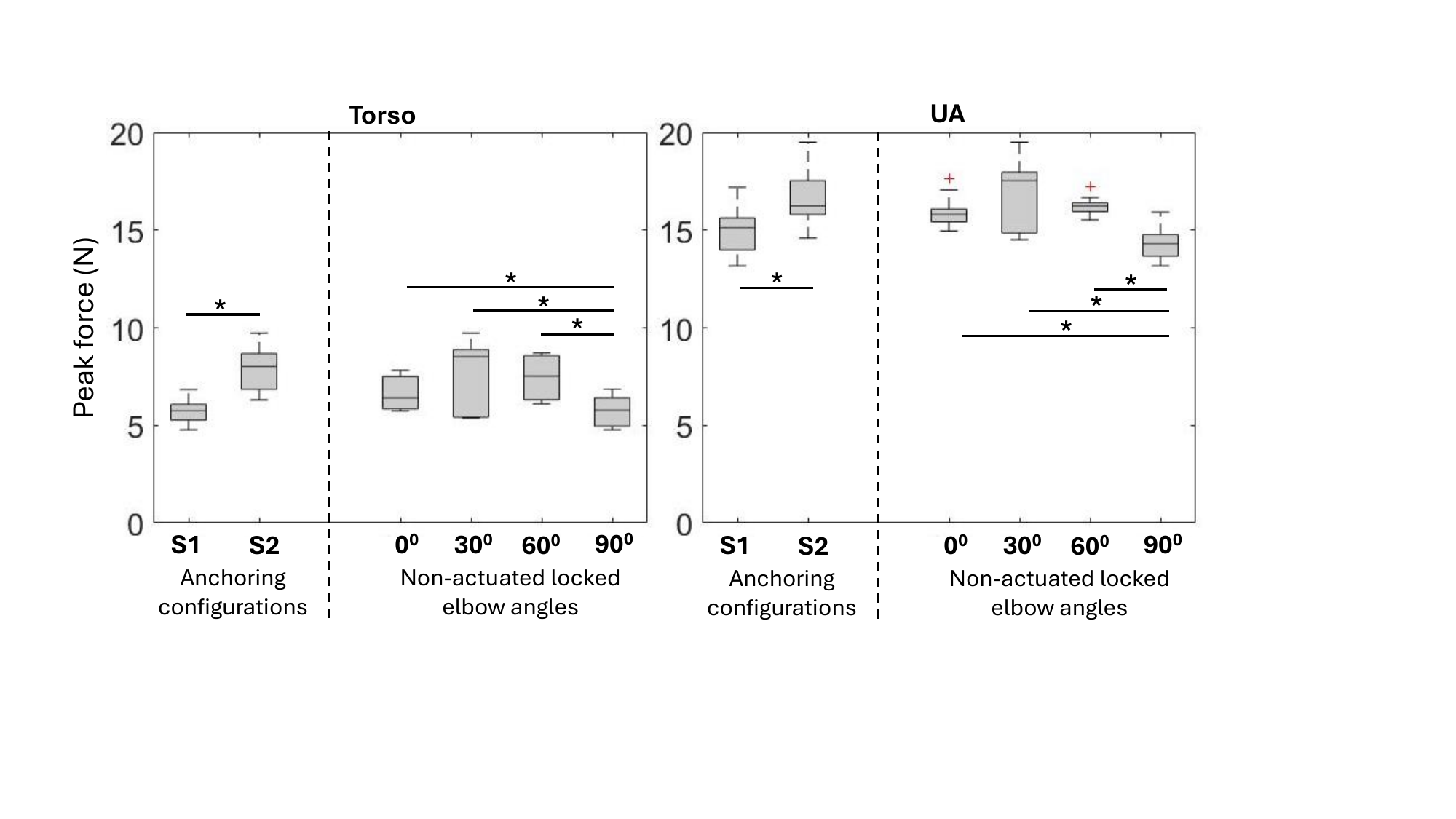}
     \vspace{-30pt}
         \caption{Boxplots of the exerted peak force for the shoulder actuator on the torso and UA. Results show significant performance differences owing to different actuator configurations and non-actuated locked elbow angles.}
     \label{fig:peakforce_stat_shoulder}
     \vspace{-18pt}
\end{figure}

\begin{figure}[!t]
\vspace{6pt}
     \centering
     \includegraphics[width=0.95\columnwidth]{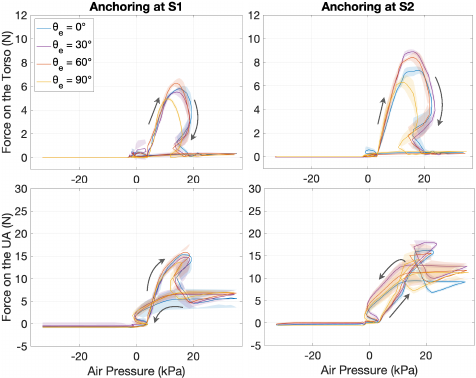}
     \vspace{-6pt}
         \caption{Evolution of the shoulder actuator forces exerted on the torso (Top) and UA (Bottom) as the internal air pressure of the actuator varies, at different conditions. Arrows demonstrate the inflation-deflation cycle. }
     \label{fig:force vs. pressure}
     \vspace{-9pt}
\end{figure}

The temporal evolution of the forces exerted by the shoulder actuator on the torso and UA across all pertinent configurations is illustrated in Fig.~\ref{fig:force vs. time}. 
When the actuator was anchored at S1, the peak force was exerted on the torso at $2.19\pm0.15$ sec and on the UA at $2.6\pm0.18$ sec. 
When the actuator was anchored at S2, the peak force was exerted on the torso at $2.27\pm0.15$ sec and on the UA at $3.07\pm0.25$ sec. Statistical analysis showed a greater peak force exerted on the torso and UA in the S2 configuration compared to S1, regardless of the non-actuated elbow locked angle ($U\textsubscript{Torso}=13997.500$, $p< 0.001$; $U\textsubscript{UA}=11512.500$, $p< 0.001$); Fig.~\ref{fig:peakforce_stat_shoulder}).
In addition, the peak force exerted on the torso and UA was significantly different across the non-actuated elbow joint locked angles ($\chi\textsuperscript{2}\textsubscript{Torso}(3)=48.389$, $p<0.001$; $\chi\textsuperscript{2}\textsubscript{UA}(2)=114.489$, $p<0.001$).
Post-hoc tests showed that locking the elbow at $90^{\circ}$ leads to a lower peak force both on the torso and UA compared to the other three cases ($p<0.001$ for all comparisons; Fig.~\ref{fig:peakforce_stat_shoulder}).
Visual inspection of Fig.~\ref{fig:force vs. pressure} suggests that exerted forces appear to vary nonlinearly with the input pressure. 
When the actuator was anchored at S1, the peak forces on the UA and torso were observed at $16.73\pm1.54$ kPa and $13.45\pm1.73$ kPa, respectively.
When the actuator was anchored at S2, the peak forces on the UA and torso were observed at $20\pm2.44$ kPa and $15.1\pm2.03$ kPa, respectively. 

The previous results altogether allow for the following observations to be made. 
In configuration S1, ROM is larger and the peak force is smaller compared to S2. 
This can be explained by the fact that, when the actuator is placed farther from the shoulder joint center along the UA, the moment arm between its force application point on the UA and the shoulder joint increases. 
This helps generate more torque for the same actuator pressure. 
Similarly, when the elbow is mostly flexed (i.e. for $\theta_e=\{60^{\circ}, 90^{\circ}\}$) the moment arm between the center of mass of the FA and the shoulder joint's axis of rotation decreases. 
This reduces the mechanical work required for rotation for the same input pressure thereby increasing shoulder joint ROM. 
Lastly, it is worth noting that when the single-cell actuator is mounted onto the rig, it is folded into two sections that behave as separate cells initially. 
During inflation, these sections push outward, increasing force on both the torso and UA. 
Around mid-inflation, they merge into a single cell causing a discontinuity in the exerted force as well as a reduction in contact with the areas covered by load cells in both the torso (mainly) and UA. 
This effect may have led to the observed nonlinearities. 

\subsection{Elbow Actuator Assessment}
\begin{figure}[!t]
\vspace{6pt}
     \centering
     \includegraphics[trim={0cm, 0cm, 0cm, 0.01cm},clip,width=0.95\columnwidth]{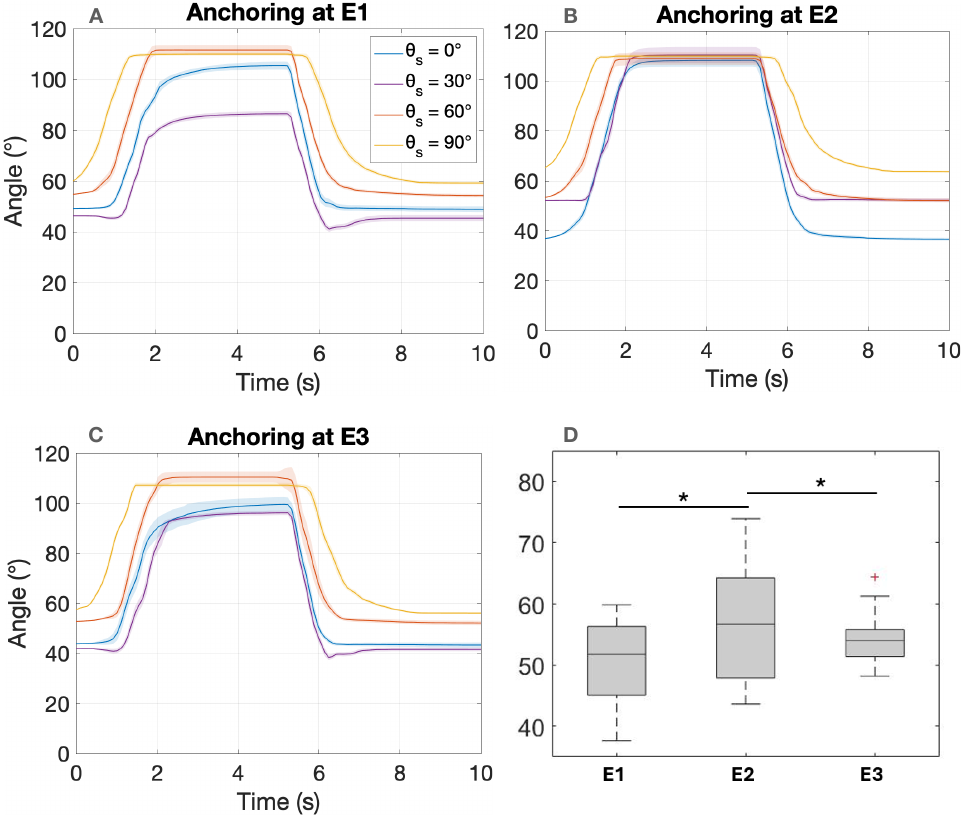}
     \vspace{-8pt}
         \caption{(A, B, C) Changes in elbow joint angle over time in the three elbow actuator anchorings for different locked shoulder angles. (D) Boxplot showing elbow joint angles across the E1, E2, and E3 configurations.}
     \label{fig:angle vs. time_elbow}
     \vspace{-14pt}
\end{figure}

The change in elbow joint angle throughout the trial and across the different non-actuated shoulder joint locked angles for all elbow actuator anchoring configurations is depicted in Fig.~\ref{fig:angle vs. time_elbow} (A-C). 
Statistical analysis revealed that the elbow joint ROM was significantly different across the elbow actuator anchoring configurations ($\chi\textsuperscript{2}(3)=28.580$, $p<0.001$; Fig.~\ref{fig:angle vs. time_elbow}D), with a greater ROM observed for E2 compared to E1 ($p<0.001$) and E3 ($p=0.012$).
There was a statistically significant effect of non-actuated shoulder joint locked angles on the elbow joint ROM for the three elbow actuator anchoring configurations (E1: $\chi\textsuperscript{2}(3)=101.544$, $p<0.001$; E2: $\chi\textsuperscript{2}(3)=102.782$, $p<0.001$; E3: $\chi\textsuperscript{2}(3)=83.809$, $p<0.001$). 
Post-hoc tests showed that the smallest elbow joint ROM was observed at a locked shoulder joint angle of $90^{\circ}$ for the E2 and E3 elbow actuator anchoring configurations, but not for E1 which was at a locked joint angle of $30^{\circ}$ ($p < 0.001$ for all comparisons).

\begin{figure}[!t]
\vspace{6pt}
     \centering
     \includegraphics[width=1\columnwidth]{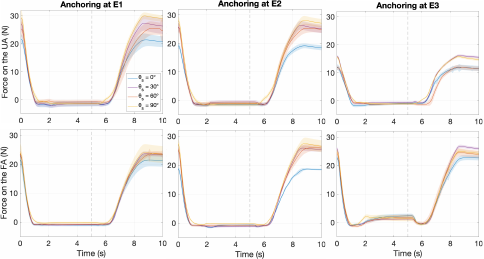}
     \vspace{-20pt}
         \caption{Temporal evolution of the forces exerted by the elbow actuator on the UA (Top) and FA (Bottom) in all pertinent configurations.}
     \label{fig:force vs. time_elbow}
     \vspace{-15pt}
\end{figure}

\begin{figure}[!th]
\vspace{3pt}
     \centering
\includegraphics[trim={1.5cm, 2cm, 6cm, 2cm},clip, width=0.95\columnwidth]{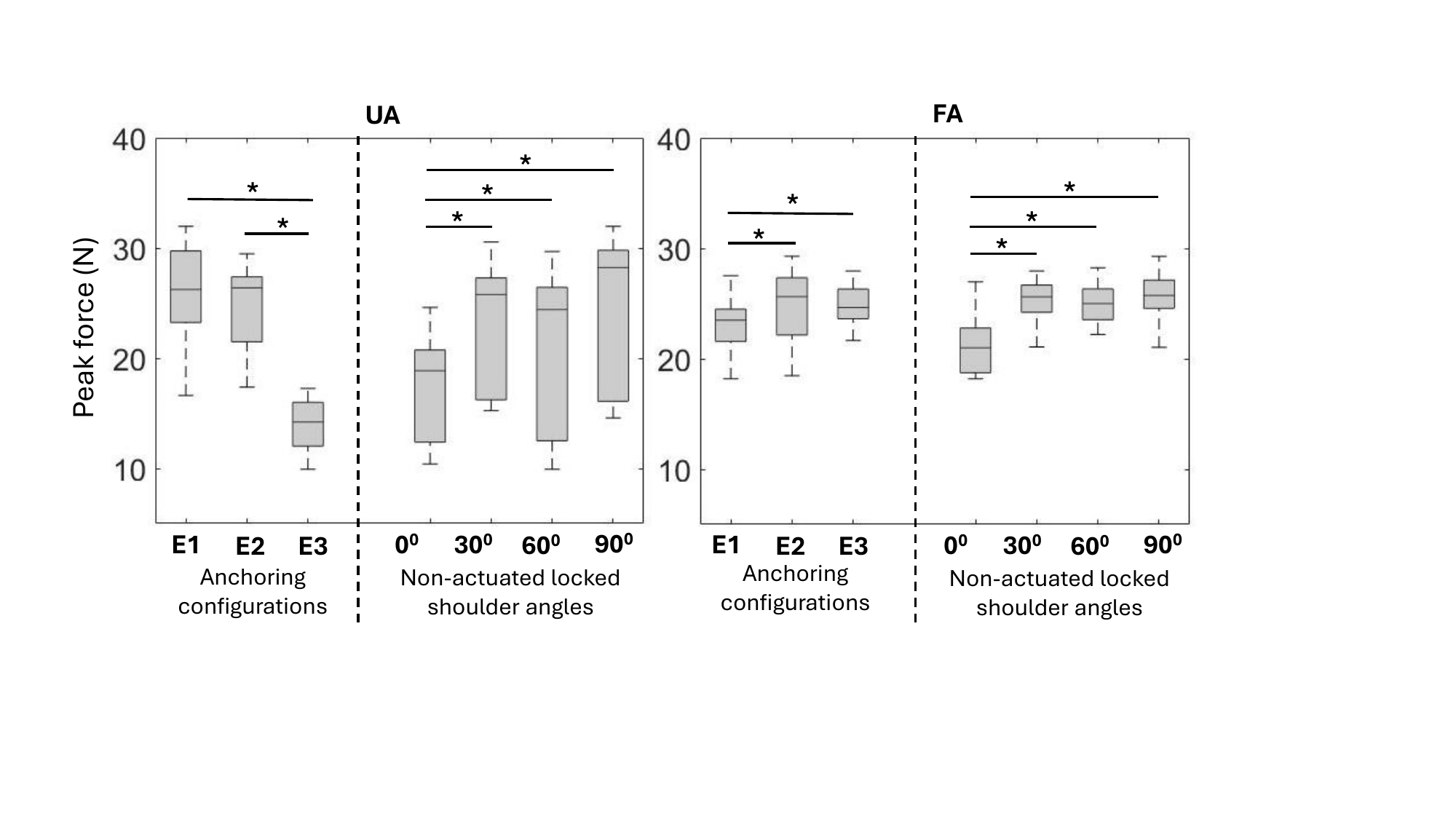}
     \vspace{-30pt}
         \caption{Boxplots of the exerted peak force for the elbow actuator on the UA and FA. Results show significant performance differences owing to different actuator configurations and non-actuated locked shoulder angles.}
     \label{fig:peakforce_stat_elbow}
     \vspace{-8pt}
\end{figure}

\begin{figure}[!t]
\vspace{6pt}
     \centering
     \includegraphics[width=1\columnwidth]{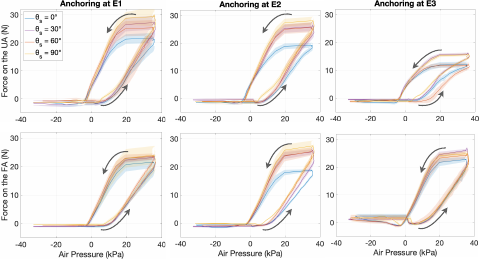}
     \vspace{-18pt}
         \caption{Evolution of the elbow actuator forces exerted on the UA (Top) and FA (Bottom) as the internal air pressure of the actuator varies. Arrows indicate deflation (flexion) and inflation (extension).}
     \label{fig:force vs. pressure_elbow}
     \vspace{-15pt}
\end{figure}

The temporal evolution of the forces exerted by the elbow actuator on the UA and FA for the three elbow actuator configurations and across the different non-actuated shoulder joint locked angles is illustrated in Fig.~\ref{fig:force vs. time_elbow}.
When the actuator was anchored at E1, the peak forces were exerted on the UA and FA at $8.51\pm0.29$ sec and $8.66\pm0.34$ sec, respectively. 
When anchored at E2, the peak forces were exerted on the UA and FA at $8.45\pm0.26$ sec and $8.53\pm0.28$ sec, respectively. 
In the E3 configuration, the peak forces were exerted on the UA and FA at $8.29\pm0.21$ sec and $8.45\pm0.32$ sec, respectively. 
Statistical analysis revealed that the peak force exerted on the UA and the FA was significantly different across the actuator anchoring configurations ($\chi\textsuperscript{2}\textsubscript{UA}(2)=243.053$, $p<0.001$; $\chi\textsuperscript{2}\textsubscript{FA}(2)=36.783$, $p<0.001$; Fig.~\ref{fig:peakforce_stat_elbow}).
Post-hoc tests showed that the force exerted on the UA was significantly lower in the E3 configuration compared to the E1 ($p<0.001$) and E2 ($p<0.001$) ones, whereas, the force exerted on the FA was significantly lower for the E1 compared to the E2 ($p<0.001$) and E3 ($p<0.001$) configurations.
In addition, the peak force exerted on the UA and FA was significantly different across the non-actuated shoulder joint locked angles ($\chi\textsuperscript{2}\textsubscript{UA}(2)=71.578$, $p<0.001$; $\chi\textsuperscript{2}\textsubscript{FA}(2)=152.088$, $p<0.001$; Fig.~\ref{fig:peakforce_stat_elbow}).
Post-hoc tests showed that locking the shoulder at $\theta_s=0^{\circ}$ led to the smallest peak force exerted on both UA and FA, compared to the other three cases ($p<0.001$ for all comparisons). 
Visual inspection of Fig.~\ref{fig:force vs. pressure_elbow} suggests that exerted forces demonstrate hysteresis to input pressure changes. 
Peak forces were observed at an average of $35.91\pm0.72$ kPa for the UA and $35.72\pm0.65$ kPa for the FA when the actuator was anchored at E1. 
When anchored at E2, the peak force was observed at $36.16\pm0.68$ kPa for the UA and $35.93\pm0.65$ kPa for the FA. 
Lastly, peak forces were observed at an average of $36.36\pm0.7$ kPa and $36.19\pm0.71$ kPa for the UA and FA, respectively, when the actuator anchored at E3.


The aforementioned results allow the following observations to be made. 
The symmetric configuration (E2) is the most efficient in terms of maximizing ROM. 
This can be explained by the mechanics of this actuator. 
When it is mounted between the UA and FA and inflated, it bends and attains an arc-like form. 
The bottom parts of its cells (i.e. the side closer to the body) come closer together while their top sides are being pulled away. 
Symmetry in the distribution of cells before and after the elbow joint center (i.e. number of cells over the UA and FA, respectively) enable the actuator to attain a more circular arc (compared to a more elliptical one when asymmetry is introduced), generating higher perpendicular force on the FA, increasing moment, and thus resulting in larger ROM. 
The asymmetric configurations bring some parts of the actuator closer together than others, reducing overall ROM but also reducing the peak force on the UA (E3; fewer cells over the UA) and FA (E1; fewer cells over the FA). 
The observed hysteresis may be caused by the binary (on/off) valves used in the pneumatic system; continuous flow valve could help remedy this effect. 
Besides the observed hysteresis, input pressure and exerted forces vary proportionally, while the transition between the deflation and inflation phases is smooth, highlighting the suitability of the elbow actuator in achieving its intended task.

\section{Conclusion} 
We investigated the contact forces generated by a single-cell actuator for shoulder abduction/adduction and a 10-cell bellow-type actuator for elbow flexion/extension using an infant-scale engineered apparatus. 
Extensive testing under various conditions revealed key information regarding actuator functionality. 
For the shoulder actuator: (1a) shifting it closer to the joint center can increase peak force while decreasing ROM, and (1b) increasing the elbow joint angle can help reduce the peak exerted force while improving ROM. 
For the elbow actuator: (2a) symmetric positioning can maximize ROM and result in equal peak forces exerted on the UA and FA, and (2b) larger shoulder joint angles can cause higher exerted forces and simultaneously reduce ROM. 

The actuator anchoring findings (1a, 2a) give insight into how to improve the design for functionality and efficiency. 
The findings relating to the effects of the adjacent joint when locked in a particular configuration (1b, 2b) can help create feedforward terms for an exosuit controller. 
For instance, if the task is to support primarily elbow function in terms of ROM, the underlying (admittance) controller should provide some input to the user to ensure the shoulder remains in a neutral position. 
A detailed treatise of such control design is part of future work.
Further, the observed nonlinearity and hysteresis between actuator input pressure and exerted forces highlight the need for a dynamics-based analysis and modeling which is another direction of future work. 

\bibliography{ipsita, Kokkoni, caio}
\bibliographystyle{IEEEtran}
\end{document}